\documentclass{article} % For LaTeX2e
\usepackage{colm2024_conference}
\colmfinalcopy

\usepackage{url}
\usepackage{hyperref}
\usepackage{enumitem}
\usepackage{microtype}

\usepackage{amsmath}
\usepackage{amssymb}

\usepackage{bm}
\usepackage{xcolor}
\definecolor{darkred}{rgb}{0.65, 0.11, 0}
\definecolor{softred}{RGB}{250,100,100}
\definecolor{softgreen}{RGB}{56,118,29}
\definecolor{softblue}{RGB}{100,150,200}

\usepackage{multicol}
\usepackage{multirow}
\usepackage{booktabs}

\usepackage{xspace}
\usepackage{graphicx}
\usepackage{pgfplots}
\usepackage{subcaption}
\pgfplotsset{compat=1.17}
\newcommand{\ourmethod}{\textsc{DAC}\xspace}

\title{DAC: Decomposed Automation Correction for Text-to-SQL}

\author{Dingzirui Wang, Longxu Dou, Xuanliang Zhang, Qingfu Zhu, Wanxiang Che \\
Research Center for Social Computing and Information Retrieval \\
Harbin Institute of Technology \\
\texttt{\{dzrwang, lxdou, xuanliangzhang, qfzhu, car\}@ir.hit.edu.cn} \\
}

\begin{document}
    \maketitle
    
        \begin{abstract}
        % text-to-SQL是一个重要的任务，通过自动合成SQL来帮助人们从数据库中获取信息
        Text-to-SQL is an important task that helps people obtain information from databases by automatically generating SQL queries.
        % 目前，基于LLM的方法是text-to-SQL任务的主流方法，而其中自动纠错方法是一种有效增强性能的方法
        Considering the brilliant performance, approaches based on Large Language Models (LLMs) become the mainstream for text-to-SQL.
        Among these approaches, automated correction is an effective approach that further enhances performance by correcting the mistakes in the generated results.
        % 然而，现有的方法都要求模型直接针对原始任务纠错，但前人研究表明模型很难正确地进行纠错，因为他们不知道如何去发现错误
        The existing correction methods require LLMs to directly correct with generated SQL, while previous research shows that LLMs do not know how to detect mistakes, leading to poor performance.
        % 因此，在本文中，我们提出从子任务的角度进行纠错，来增强text-to-SQL的性能
        Therefore, in this paper, we propose to employ the decomposed correction to enhance text-to-SQL performance. 
        % 我们首先从理论上论证了对于任意任务，从子任务的角度进行纠错的性能要优于直接对原始任务进行纠错
        We first demonstrate that decomposed correction outperforms direct correction since detecting and fixing mistakes with the results of the decomposed sub-tasks is easier than with SQL. 
        % 基于上述分析，我们提出\ourmethod，纠错时首先生成SQL对应实体和骨架，然后比较初始SQL和生成的实体和骨架之间的差异，作为反馈进行纠错
        Based on this analysis, we introduce Decomposed Automation Correction (\ourmethod), which corrects SQL by decomposing text-to-SQL into entity linking and skeleton parsing.
        \ourmethod first generates the entity and skeleton corresponding to the question and then compares the differences between the initial SQL and the generated entities and skeleton as feedback for correction.
        % 实验结果表明，我们的方法在Spider、KaggleDBQA和Bird三个数据集上，相对baseline带来了$\%$的性能提升，证明了我们方法的有效性
        Experimental results show that our method improves performance by $3.7\%$ on average of Spider, Bird, and KaggleDBQA compared with the baseline method, demonstrating the effectiveness of \ourmethod\footnote{Our code and data are released in \url{https://github.com/zirui-HIT/DAC}.}.
    \end{abstract}

    \section{Introduction}
        % Text-to-SQL是一个重要的任务，通过基于用户问题和数据库生成相应的SQL，来显著降低从数据库中获取信息的难度
Text-to-SQL is an important task, significantly reducing the overhead of obtaining information from the database by generating corresponding SQL based on user questions \citep{Text2SQL-Survey-PLM}.
% 具体来说，text-to-SQL任务定义为，输入用户问题和相关的数据库信息，返回相应的SQL结果
Specifically, about the text-to-SQL task, users provide the question and the related database, and then the model generates the corresponding SQL results.
% 目前，基于LLM的方法已经成为text-to-SQL领域的主流方法，因为其无需微调就能取得超过微调模型的性能
Currently, the methods based on Large Language Models (LLMs) have become the mainstream for text-to-SQL because LLMs can achieve higher performance than fine-tuned models, without fine-tuning \citep{Text2SQL-Survey-LLM,li2024texttosqlsurvey,shi2024texttosqlsurvey,kanburoglu2024texttosqlsurvey}.
% 因此，本文主要关注如何基于大模型解决text-to-SQL任务
Therefore, in this paper, we mainly focus on how to enhance the performance of the text-to-SQL task based on LLMs.

% 在现有的研究中，自纠错是一种有效提升text-to-SQL的方法，即LLM会首先生成一个初始的SQL，然后再判断SQL中的错误之处并加以改正
Recent research shows that automated correction is an effective method to improve text-to-SQL performance \citep{Self-Debug,MAGIC,xie2024deasql,MAC-SQL}, where LLMs first generate an initial SQL and then detect mistakes of the SQL as correction feedback \citep{Self-Debug-Survey}. 
% 例如，SQL-CRAFT会基于SQL生成的结果，来判断生成结果是否满足用户问题，并据此对SQL进行改正
For instance, SQL-CRAFT~\citep{SQLCraft} assesses whether the SQL execution results on databases satisfy the user question and corrects the SQL accordingly.
% 而EPI-SQL会构建错例的示例库，每次纠错时会从中检索相似的示例，来引导生成正确的纠错结果
EPI-SQL~\cite{EPI-SQL} builds an example library of error cases and then retrieves similar cases from the library to guide each correction.
% 然而前人工作表明，现有的LLM并不能很好地发现SQL中的错误并进行纠错，因为模型会对自己生成的结果很自信
However, previous studies indicate that it could be not effective for LLMs to directly correct all mistakes in the generated SQL since it is hard for LLMs to detect mistakes in the generated results \citep{Benchmarking,kamoi2024SelfCorrectionSurvey,zhang2024selfcontrast}.

\begin{figure}[t]
    \centering
    \includegraphics[width=0.7\linewidth]{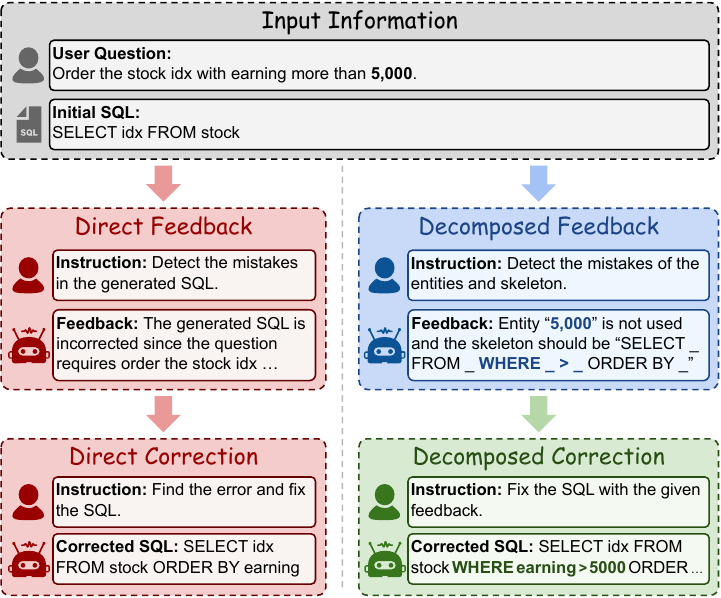}
    \caption{
        The comparison between direct correction and decomposed correction.
        Direct correction shows poor performance since LLMs do not know how to detect mistakes.
        Decomposed correction brings better performance since it is easier to detect mistakes from the decomposed tasks of entity linking and skeleton parsing.
    }
    \label{fig:motivation}
\end{figure}

% 前人工作表明，针对推理过程进行纠错有比直接对原始任务的结果纠错更好的性能，因为推理过程要比最终结果更容易发现错误
Previous works \cite{dhuliawala2024cove,vacareanu2024general} show that correction with the decomposed reasoning process has better performance than directly correcting because detecting and correcting the mistakes in the reasoning process is easier than in the final answer. 
% 因此在本文中，我们讨论：
Therefore, in this paper, we discuss that:
% 1. 理论上，论证了将原始任务分解为子任务后，根据子任务的结果进行纠错，要比直接对原始任务的结果纠错更有效
\textit{(i)} Experimentally, we present that \textbf{\textit{correction with decomposed sub-tasks is more effective than direct correction}};
% 2. 方法上，我们提出将text-to-SQL任务分解为实体链接和骨架生成两个子任务，来增强自纠错方法的性能
\textit{(ii)} Methodologically, we propose \textbf{\textit{correcting the text-to-SQL results based on the decomposed tasks of entity linking and skeleton parsing}} to enhance the performance of the automation correction.

% 首先，我们从理论上分析了，将原始任务分解为子任务进行纠错，会有更好的纠错性能
First, we discuss that correction with the decomposed sub-tasks can bring better correction performance than the direct correction, as shown in Figure~\ref{fig:motivation}. 
% 我们首先讨论了对于任意任务，分解为子任务之后纠错，从子任务的角度能更好地发现错误，从而有比基于原始任务纠错更好的纠错性能
We first present that decomposing into sub-tasks ensures better automation correction performance since mistakes can be more easily detected and corrected based on sub-tasks than using the original generation result.
% 将上述结论应用到text-to-SQL任务，我们提出将text-to-SQL分解为实体链接和实体生成两个子任务，他们是解决text-to-SQL任务最重要的两个能力
Applying the above conclusion, we propose decomposing text-to-SQL into two sub-tasks: entity linking and skeleton parsing, which are the two most crucial capabilities for addressing the text-to-SQL task \cite{RESDSQL,Text2SQL-Survey-PLM,Text2SQL-Survey-LLM}.

% 基于上述分析，我们提出我们的方法，通过预生成实体和骨架，然后将不一致的实体或骨架作为纠错信息
Based on the above analysis, we propose \textbf{D}ecomposed \textbf{A}utomation \textbf{C}orrection (\ourmethod), which corrects the generated SQL with the mistakes of entity linking and skeleton parsing as feedback.
% 为了判断问题相关的数据库实体，根据前人工作，我们使用LLM来根据用户问题和数据库直接生成相关实体
Schema linking denotes detecting the table and column names related to the question, where we use LLMs to directly generate relevant entities based on the user question and the database, following \citet{CodeIE}. 
% 为了判断SQL骨架，依据前人工作，我们使用LLM来猜测问题对应的SQL骨架
Skeleton parsing denotes generating the SQL skeleton corresponding to the user question, where we employ LLMs to generate the SQL skeleton according to the prior research \cite{DESEM,RetrievalAugment}.
% 在生成完实体和骨架后，将他们和初始生成的SQL的不一致之处作为反馈，来纠正SQL，如图所示
After generating the entities and skeleton, we take their inconsistencies with the initial SQL as feedback for the correction, as shown in Figure~\ref{fig:motivation}.

% 为了验证我们的方法，我们在Spider、Bird和KaggleDBQA三个主流text-to-SQL数据集上进行了实验
To validate \ourmethod, we conduct experiments on three mainstream text-to-SQL datasets: Spider~\cite{Spider}, Bird~\cite{Bird}, and KaggleDBQA~\cite{KaggleDBQA}.
% 实验表明，我们的方法相对不使用self-debug平均提升了3.7%，并且相对前人的text-to-SQL纠错方法提升了1.3%，证明了我们方法有效性
The experiments show that our method achieves an average improvement of $3.7\%$ compared to baselines, and a $1.3\%$ improvement over previous text-to-SQL automation correction methods, demonstrating the effectiveness of \ourmethod.

% 我们的贡献如下：
Our contributions are as follows:
\begin{itemize}% [nolistsep,leftmargin=*]
    % 我们提出了一套分析纠错性能的理论，论证了从细粒度的角度提供纠错反馈可以增强纠错的性能
    \item We experimentally discuss that decomposed correction has better performance than direct correction;
    % 我们提出了我们的方法，通过提供更细粒度的自纠错信息来增强text-to-SQL自纠错性能
    \item We present \ourmethod, which improves text-to-SQL automated correction performance based on the decomposed sub-tasks of entity linking and skeleton parsing;
    % 实验表明，我们的方法相对前人最好的text-to-SQL自纠错方法提升了1.3，证明了我们方法的有效性
    \item Experiments show that \ourmethod improves performance by $3.7\%$ compared to baselines and $1.3\%$ compared to the previous text-to-SQL automated correction methods, proving the effectiveness of our method.
\end{itemize}

    \section{Discussion}
        \begin{figure}[t]
    \centering
    \includegraphics[width=0.9\linewidth]{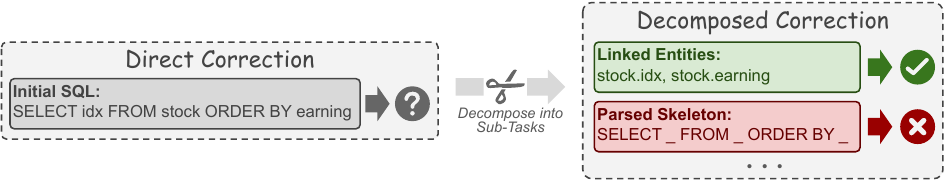}
    \caption{
        % “针对子任务结果纠错要比针对SQL纠错的性能要好”的示意图，以问题“Order the stock idx with earning more than 5,000”为例
        The illustration of our discussion, taking the question ``\textit{Order the stock idx with earnings more than 5,000''} as an example.
        % 考虑完整的SQL时，模型很难找出具体的错误之处
        About the direct correction (left part), it is challenging for LLMs to pinpoint specific mistakes.
        % 而对SQL进行分解后，模型能更容易地判断出错误之处，并正确地修改相应的错误
        After decomposing the SQL into sub-tasks (right part), it is easier for LLMs to identify and correct the mistakes.
    }
    \label{fig:analysis}
\end{figure}

% 在这一节，我们讨论：将text-to-SQL分解为子任务进行纠错，能有效增强SQL纠错性能
In this section, we discuss that \textbf{decomposing text-to-SQL into sub-tasks can effectively enhance SQL correction performance}.
% 首先我们论证：对于任意任务，将其分解为子任务进行纠错可以增强在各个子任务上的纠错性能，从而增强纠错的性能
First, we demonstrate that correction by decomposing into sub-tasks is more effective than direct correction since it is easier for LLMs to detect and correct mistakes based on the sub-tasks.
% 然后，我们针对text-to-SQL任务，提出将其分解为实体链接和骨架生成两个子任务进行纠错，增强SQL纠错的性能
Then, we propose to correct the text-to-SQL results by decomposing the task into two sub-tasks: entity linking and skeleton parsing, thereby enhancing the SQL correction performance.

\subsection{Performance of Sub-Task Correction}
    % 我们将分解子任务进行纠错的正确性看作两部分：纠错结果符合子任务，符合子任务的纠错结果也符合原始任务
    We consider the correction performance of task decomposition in two parts: correction results satisfy the sub-tasks, and correction results that satisfy the sub-tasks also satisfy the original task.
    % 我们提出：对于任意任务，将任务分解为子任务之后进行纠错，纠错的结果能更好地符合各个子任务，但是符合子任务的纠错结果不一定也符合原始任务
    We propose that decomposing the task into sub-tasks can make the correction results better satisfy the sub-tasks, while the correction results satisfying the sub-tasks could not satisfy the original task.
    % 上述结论的示意图如图所示
    The illustration of the above proposition is shown in Figure~\ref{fig:analysis}.

    % 直观来看，原始任务可以被看成是多个子任务的组合
    Intuitively, the original task can be viewed as a combination of multiple sub-tasks.
    % 由于我们解决的子任务被原始任务包含，根据前人工作，解决这些子任务的难度要比解决原始任务要简单
    Since the sub-tasks used are included in the original task, according to \citet{wang2024bridge}, generating correct answers to sub-tasks is easier than the original task, thereby it is easier to detect the mistakes based on these sub-tasks than the original task.
    % 同样因为选取的子任务无法完全覆盖原始任务，导致即使纠错的结果能满足所有的子任务，也无法保证结果完全满足原始任务
    However, because the used sub-tasks do not entirely encompass the original task, even if the correction results satisfy all the used sub-tasks, it cannot guarantee that the results fully meet the original task. 
    % 因此，分解子任务时，需要确保子任务能尽量覆盖原始任务，使得满足子任务的纠错结果尽量也能满足原始任务
    Therefore, when decomposing sub-tasks, it is necessary to ensure that the sub-tasks can maximally cover the original task so that the correction results satisfying the sub-tasks are more likely to satisfy the original task.

\subsection{Adaption to Text-to-SQL}
    % 我们将上述结论应用到text-to-SQL任务上，选用实体链接和骨架生成作为子任务
    We apply the discussion above to the text-to-SQL task, where we use entity linking and skeleton parsing as sub-tasks for automated correction.
    % 实体链接指的是从给定的数据库结构中，找到和问题相关的表明和列名
    Entity linking refers to identifying relevant table and column names from the given database schema based on the user question \cite{EtA}. 
    % 骨架生成指的是基于用户问题生成相应的SQL骨架
    Skeleton parsing denotes creating the corresponding SQL skeleton of the user question, which removes the table names, column names, and values in SQL \cite{DESEM}. 
    % 我们选用这两个子任务是因为，他们被前人工作看作text-to-SQL最重要的两个能力
    We use these two sub-tasks because previous work regards them as the most crucial capabilities for text-to-SQL \cite{RESDSQL,Text2SQL-Survey-PLM,Text2SQL-Survey-LLM}.

    While based on the discussion above, correction of the sub-tasks could not ensure the result correctness of the text-to-SQL task. 
    % 因此，我们统计了实体和骨架都正确的SQL中，是正确SQL的比例
    Therefore, we statistic the proportion of the correct SQL in which the entity and skeleton are correct.
    % 统计结果如表所示，都非常接近百分之百，证明了我们分解子任务进行纠错的方法有效性
    The statistical results, as shown in Table~\ref{tab:result_accuracy}, are very close to one hundred percent
    % 这一结果表明符合我们提出的子任务的纠错结果也基本符合text-to-SQL任务，证明了我们分解子任务进行纠错的方法有效性
    This result shows that the correction results in line with our proposed sub-tasks are also basically in line with the text-to-SQL task, demonstrating the effectiveness of our sub-task decomposition method for the text-to-SQL automated correction.

    \section{Methodology}
        \begin{figure*}[t]
    \centering
    \includegraphics[width=\textwidth]{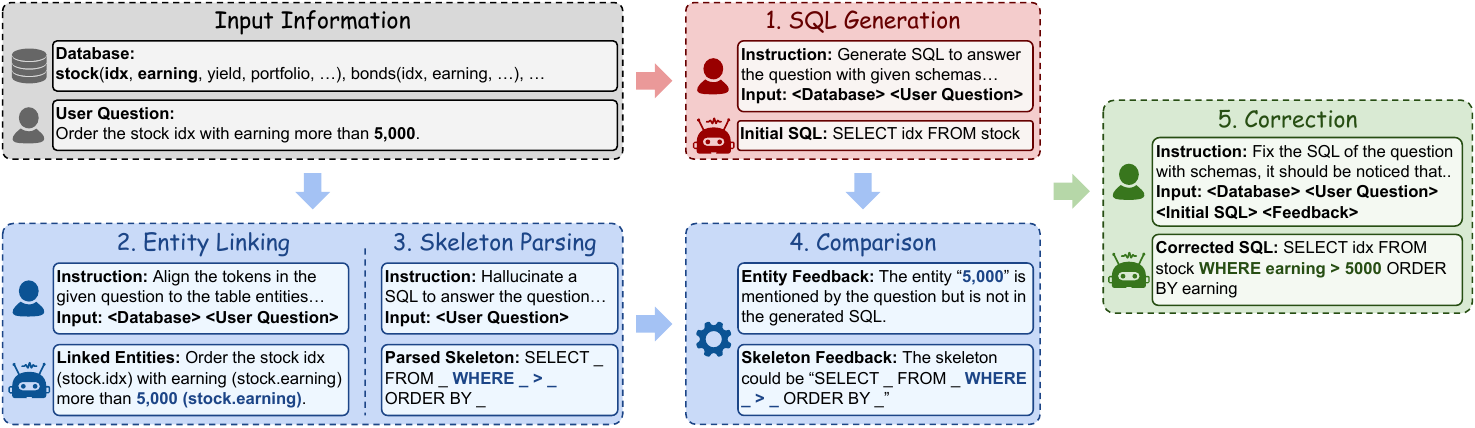}
    \caption{
        The pipeline of \ourmethod, which consists of five steps:
        \textbf{\textit{(i)} SQL Generation}: Generate the initial SQL; 
        \textbf{\textit{(ii)} Entity Linking}: Detect the question-related entities; 
        \textbf{\textit{(iii)} Skeleton Parsing}: Generate the SQL skeleton of the question;
        % 判断初始SQL和生成的实体、骨架的不一致之处，作为反馈
        \textbf{\textit{(iv)} Comparison}: Determine the inconsistencies between the initial SQL and the linked entities and parsed skeletons as feedback;
        \textbf{\textit{(v)} Correction}: Correct the SQL with the comparison feedback.
    }
    \label{fig:pipeline}
\end{figure*}

% 在这一部分，我们介绍我们的方法，从实体和骨架两个角度来生成错误信息，作为反馈进行自纠错
Following the discussion above, in this section, we present \ourmethod, which generates automated correction feedback by decomposing the text-to-SQL task into the sub-tasks of entity linking and skeleton parsing.
% 我们方法的示意图如图所示
The illustration of \ourmethod is shown in Figure~\ref{fig:pipeline}, and the prompts of \ourmethod are present in the supplementary material.

\subsection{SQL Generation}
    % 在纠错之前，我们需要先成一个被纠错的SQL，作为初始的SQL
    Before the correction, we first generate an initial SQL to be corrected.
    % 遵循前人工作，我们使用few-shot来生成SQL
    Following previous work \citep{SQLCraft}, we use few-shot learning to generate SQL.
    % 我们使用BM25从给定的示例池中选取和用户问题相似的示例，和数据库、用户问题一起作为输入
    We use BM25 to select demonstrations similar to the user question from the given demonstration pool, together with the database and the user question as the input.
    % 输出时，模型直接输出单个SQL作为初始的结果
    With the input, LLMs directly generate a single SQL as the initial result to be corrected.

\subsection{Entity Linking}
    % 在获得SQL后，我们希望知道它在实体连接上的错误
    After obtaining the initial SQL, we aim to identify mistakes in the linked entities. 
    % 为此，我们需要知道正确的问题相关的实体，即问题相关的数据库中的表和列
    To achieve this, it is essential to know the entities related to the user question, specifically the table and column names in the question-related database. 
    % 遵循前人工作，我们使用LLM来选取问题相关的表和列
    Following previous work \citep{andrew2024entityextraction,goel2023informationextraction}, we use LLMs to select the question-related tables and columns. 
    % 我们使用分词后的用户问题以及数据库作为输入，python中List[Dict[str, str]]格式的实体连接信息作为输出，长度与分词后的问题中的token数一致
    We employ the tokenized user question and the database as inputs and output the entity linking information in the format of \texttt{List[Dict[str, str]]} in Python, where the length of the list matches the number of tokens in the tokenized question. 
    % 我们使用这种格式，是因为LLM生成程序格式的实体连接的性能要好于直接抽取实体
    We use this format because the performance of entity linking generated in programmatic format by LLMs is superior to only generating entities \citep{CodeIE}.

    % 对于每个Dict，参照\cite，它分别包括三个key：
    Following \citet{EtA}, each \texttt{Dict} includes three keys:
    % (i) token：在原始问题中对应的token
    \textit{(i) token}: the corresponding token in the original question;
    % (ii) schema：在数据库中对应的表名或列名
    \textit{(ii) schema}: the corresponding table name or column name in the database;
    % (iii) type：包括tab、col、val，分别表示该token对应表名、列名或一个条件值
    \textit{(iii) type}: including ``\textit{tab}'', ``\textit{col}'', and ``\textit{val}'', representing the table name, column name, or a condition value that the token corresponds to respectively.

\subsection{Skeleton Parsing}
    % 在得知正确的实体连接后，我们还希望知道正确的骨架信息，来判断初始SQL生成骨架上的错误
    After identifying the linked entities, we seek to parse the skeleton corresponding to the user question to identify mistakes in the initial SQL generation on the skeleton. 
    % 前人研究表明，LLM能够只通过用户问题猜测出SQL对应的骨架，因为大部分时候骨架对应了问题的语法逻辑，无需输入数据库
    Previous research indicates that LLMs can parse the SQL skeleton based solely on the user question since the skeleton corresponds to the syntactic logic of the question without requiring database information \citep{ZeroNL2SQL,SQL-Encoder,kothyari2023crush}.
    % 因此，我们输入用户问题，让LLM猜测出一个问题对应的SQL
    Therefore, we use only the user question as the input and ask LLMs to parse an SQL skeleton corresponding to the question.
    % 然后，我们剔除掉SQL中的实体，因为这些实体可能无法与原始数据库中的实体对应，从而得到SQL的骨架
    Subsequently, we remove entities from the parsed SQL, since these entities could not correspond to the column names or table names in the original database, thus obtaining the SQL skeleton.

\subsection{Comparison}
    % 在得到正确的实体连接和骨架后，我们直接比较差异判断初始SQL中的错误部分，而无需使用LLM
    After obtaining the linked entities and parsed skeleton, we directly compare and identify the inconsistency parts between them and the initial SQL without LLMs as the correction feedback.
    % 我们首先按照SQL语法抽取出初始SQL中的实体和骨架
    We first extract the entities and skeletons in the initial SQL according to the SQL syntax.
    % 对于实体连接，我们判断哪些实体被问题提及却没有在SQL中被使用，来作为输入
    For \textit{linked entities}, we determine which linked entities are mentioned in the question but not used in the initial SQL. 
    % 我们并没有将SQL包含但问题没有提及的实体作为错误，因为某些SQL用到的实体确实可能没有被用户问题提及
    We do not consider entities that are present in the SQL but not mentioned in the question as mistakes because the user question could not explicitly mention some entities used in the SQL.
    % 例如，有的表可能用来bridge另外两个表，而没有被用户问题提及
    For example, some tables are used to bridge the other two tables by the foreign keys without being mentioned by the user question.
    % 对于骨架，我们对那些初始SQL骨架与预测的骨架不一致的数据进行纠错，并且直接输入预测的骨架
    About the \textit{parsed skeleton}, we correct the initial SQLs whose skeletons are inconsistent with the parsed skeletons. 
    % 我们并没有细致地告诉模型具体缺少或增多了骨架中的哪些关键词，因为不同的关键词在SQL中可能在多个地方多次出现，太过具体的信息可能会让模型不知道具体该修改哪个
    We input the completed parse skeleton as the feedback, where do not indicate specific mistaken keywords because different keywords could appear multiple times in the SQL, while too specific information could confuse the model about which keywords exactly need to be modified.

\subsection{Correction}
    % 在得到错误信息后，我们使用其对初始SQL进行纠错
    Upon constructing correction feedback, we use them to correct the initial SQL.
    % 如果SQL执行时产生了错误信息，我们会将错误信息作为反馈来进行纠错，遵循前人工作
    If an error message occurs during SQL execution, we use the error message as feedback, following \citet{Self-Debug}.
    % 我们将数据库、用户问题、初始SQL和反馈作为输入，让模型输出纠错后的SQL
    We take the database, user question, initial SQL, and feedback as input, and ask LLMs to output the corrected SQL.
    % 纠错时，我们先对实体连接进行纠错，然后对骨架进行纠错
    During the correction, we first correct entity mistakes, followed by skeleton mistakes. 
    % 我们并没有同时纠错，是因为过于复杂的错误信息可能会让模型感到困惑
    We do not correct both simultaneously because overly complex feedback could confuse the LLM. 
    % 而先实体连接后骨架，是因为实体连接信息能帮助模型进行骨架的修复，因此先引入实体连接信息
    We first correct the linked entities since the entity information can help LLMs correct the skeleton.
    % 例如，用户问题问“排名最高的学生成绩”，如果数据库中直接提供了“排名”这个列，那么SQL可以直接选取“排名=1”的数据，而不需要“ORDER BY 成绩”
    For example, if the user question asks for the ``\textit{student with the top-ranked score}'', and the database directly provides a ``\textit{rank}'' column, the SQL can directly select data with ``\textit{rank=1}'' without ``\textit{ORDER BY score}''.

    \section{Experiments}
        \subsection{Settings}
    \subsubsection{Datasets}
        % 我们使用Spider、Bird和KaggledDBQA三个数据集来验证我们的方法
        We use Spider~\citep{Spider}, Bird~\citep{Bird}, and KaggleDBQA~\citep{KaggleDBQA} to evaluate \ourmethod.
        % Spider是主流的text-to-SQL数据集之一，数据库涵盖了多表、多领域
        Spider is one of the most mainstream text-to-SQL datasets, which covers multi-table and multi-domain questions.
        % Bird相较于Spider，问题更难，数据库更复杂，更加贴合实际应用场景
        Compared with Spider, Bird is more difficult, and databases are more complex and more near to the practical application.
        % KaggleDBQA与Bird的难度接近，但数据库的规模更加庞大（为了简略，我们将KaggleDBQA称作Kaggle）
        The difficulty of KaggleDBQA is close to that of Bird, but the scale of the database is larger\footnote{We call KaggleDBQA as \textbf{Kaggle} for simplicity.}.

    \subsubsection{Metric}
        % 在本文中，遵循前人工作，我们使用执行准确率（EX）来评价我们的方法
        In this paper, following the previous works \citep{MAGIC,SQLFixAgent}, we use execution accuracy (EX) to evaluate our method.
        % EX是指，预测的SQL的执行结果与正确答案的SQL的执行结果相同的数据的比例
        EX refers to the proportion of data that predicted SQL execution results are the same as correct SQL execution results.

    \subsubsection{Models}
        % 我们在Llama3、Deepseek-Coder的sft版本以及gpt-3.5-turbo上验证我们的方法
        We verify our methods on the instruction fine-tuning version of Llama3~\citep{Llama3}, Deepseek-Coder (Deep.C.) \citep{Deepseek-Coder}, and \texttt{gpt-3.5-turbo-1106} (GPT-3.5) \citep{gpt-3.5-turbo}.
        % Llama3和Deepseek-Coder分别是目前开源模型中，最主流的通用模型和代码模型之一
        Llama3 and Deepseek-Coder are the most mainstream general and code LLMs in the current open-source LLMs, respectively.
        % 而gpt-3.5-turbo是目前闭源模型中，最主流的模型之一
        While GPT-3.5 is one of the most popular closed-source LLMs.
        % 因此，上述的模型选择能很好地覆盖不同的应用场景
        Therefore, our experimental LLMs can well cover different application scenarios.

    \subsubsection{Implementation Details}
        % 对于我们方法中的每一步，我们使用few-shot来进行推理
        We employ few-shot inference for each step in \ourmethod.
        % 遵循前人工作，我们使用5-shot，将Spider训练集作为候选示例池，并使用\citep标注的实体连接信息，使用BM25对于每个问题选取最相似的示例
        Following the previous work \citep{ODIS}, we use 5-shot for the inference.
        We use the Spider training set as the candidate demonstration pool and use BM25 to select the most similar demonstrations for each user question.
        For the entity linking step, we employ the labeled linking information of Spider collected by \citet{EtA}, which is used to guide the generation format \cite{madaan2023makes} and can be adapted to different databases as shown in Table~\ref{tab:main_experiment}.
        The prompts we used can be seen in the supplementary material.
        % 我们使用了两张A100-80g，针对每个设置用时两个小时
        % For our experiments, we use two A100-80G which cost two hours for each setting.

\subsection{Main Experiment}
    \begin{table}[t]
        \small
        \centering
        % \begin{tabular}{lll|ccc|c}
%     \toprule
%     \textbf{Model} & \textbf{Scale} & \textbf{Method} & \textbf{Spider} & \textbf{Bird} & \textbf{Kaggle} & \bm{$\Delta$} \\
%     \midrule
%     \multirow{4}{*}{Llama3} & \multirow{2}{*}{8b} & few-shot & $70.7$ & $30.4$ & $24.6$ & \multirow{2}{*}{$+5.3$} \\
%      &  & \quad + \ourmethod & \bm{$75.0$} & \bm{$37.3$} & \bm{$29.4$} &  \\
%     \cmidrule{2-7}
%      & \multirow{2}{*}{70b} & few-shot & $80.7$ & $47.0$ & $31.4$ & \multirow{2}{*}{$+1.9$} \\
%      &  & \quad + \ourmethod & \bm{$81.3$} & \bm{$48.4$} & \bm{$35.1$} &  \\
%     \midrule
%     \multirow{4}{*}{Deep.C.} & \multirow{2}{*}{6.7b} & few-shot & $74.1$ & $42.4$ & $21.7$ & \multirow{2}{*}{$+3.0$} \\
%      &  & \quad + \ourmethod & \bm{$77.1$} & \bm{$44.7$} & \bm{$25.4$} &  \\
%     \cmidrule{2-7}
%      & \multirow{2}{*}{33b} & few-shot & $77.0$ & $48.3$ & $27.0$ & \multirow{2}{*}{$+3.2$} \\
%      &  & \quad + \ourmethod & \bm{$80.1$} & \bm{$52.7$} & \bm{$29.2$} &  \\
%     \midrule
%     \multirow{2}{*}{\texttt{gpt-3.5-turbo}} & \multirow{2}{*}{-} & few-shot & $76.9$ & $40.0$ & $24.3$ & \multirow{2}{*}{$+5.1$} \\
%      &  & \quad + \ourmethod & \bm{$80.6$} & \bm{$44.9$} & \bm{$30.9$} &  \\
%     \midrule
%     \multicolumn{3}{c|}{$\Delta$} & $+2.9$ & $+4.0$ & $+4.2$ & $+3.7$ \\
%     \bottomrule
% \end{tabular}

\begin{tabular}{ll|ccc}
    \toprule
    \textbf{Model} & \textbf{Method} & \textbf{Spider} & \textbf{Bird} & \textbf{Kaggle} \\
    \midrule
    \multirow{2}{*}{Llama3-8b} & few-shot & $70.7$ & $30.4$ & $24.6$ \\
     & \quad + \ourmethod & \bm{$75.0$} & \bm{$37.3$} & \bm{$29.4$} \\
    \midrule
    \multirow{2}{*}{Llama3-70b} & few-shot & $80.7$ & $47.0$ & $31.4$ \\
     & \quad + \ourmethod & \bm{$81.3$} & \bm{$48.4$} & \bm{$35.1$} \\
    \midrule
    \multirow{2}{*}{Deep.C.-6.7b} & few-shot & $74.1$ & $42.4$ & $21.7$ \\
     & \quad + \ourmethod & \bm{$77.1$} & \bm{$44.7$} & \bm{$25.4$} \\
    \midrule
    \multirow{2}{*}{Deep.C.-33b} & few-shot & $77.0$ & $48.3$ & $27.0$ \\
     & \quad + \ourmethod & \bm{$80.1$} & \bm{$52.7$} & \bm{$29.2$} \\
    \midrule
    \multirow{2}{*}{GPT-3.5} & few-shot & $76.9$ & $40.0$ & $24.3$ \\
     & \quad + \ourmethod & \bm{$80.6$} & \bm{$44.9$} & \bm{$30.9$} \\
    \bottomrule
\end{tabular}

        \caption{
            The main experiment results of \ourmethod.
            The baseline few-shot method uses the initial SQL as the result.
            The improved results led by \ourmethod are marked in \textbf{bold}.
        }
        \label{tab:main_experiment}
    \end{table}

    \begin{table*}[t]
        \centering
        \small
        \begin{tabular}{ll|cc}
    \toprule
    \textbf{Model} & \textbf{Method} & \textbf{Spider} & \textbf{Bird} \\
    \midrule
    \multirow{3}{*}{\texttt{gpt-3.5-turbo}} & Self-Debug~\citep{Self-Debug} & $71.1 \rightarrow 72.2$ & $-$ \\
     & SQL-CRAFT~\citep{SQLCraft} & $78.2 \rightarrow 79.3$ & $-$ \\
     & \ourmethod & $76.9 \rightarrow \bm{80.6}$ & $40.0 \rightarrow 44.9$ \\
    \midrule
    \multirow{4}{*}{\texttt{gpt-4}} & Self-Debug~\citep{Self-Debug} & $73.2 \rightarrow 73.6$ & $-$ \\
     & EPI-SQL~\citep{EPI-SQL} & $79.7 \rightarrow 85.1$ & $-$ \\
     & SQL-CRAFT~\citep{SQLCraft} & $79.7 \rightarrow 85.4$ & $53.3 \rightarrow 55.2$ \\
     & MAGIC~\citep{MAGIC} & $78.6 \rightarrow \bm{85.7}$ & $-$ \\
    \bottomrule
\end{tabular}

        \caption{
            The comparison of \ourmethod with other text-to-SQL correction methods on GPT-3.5 and \texttt{gpt-4}.
            Since the performance is affected by the API version, we report the performance change after using each method.
            We do not report the performance on KaggleDBQA since other text-to-SQL correction methods do not report the performance in their paper on this dataset.
            We do not adapt \ourmethod to \texttt{gpt-4} due to the high cost of the API.
        }
        \label{tab:comparison_with_other_system}
    \end{table*}

    % 我们方法的主实验如表所示，从表中可以发现，我们的方法在所有数据集和所有模型上，均带来了性能提升，平均提升3.7，证明了我们方法的有效性和泛用性
    The main experiment results of our method are shown in Table~\ref{tab:main_experiment}.
    From the table, it can be found that \ourmethod brings performance improvement on all datasets and all models, with an average improvement of $3.7\%$, proving the effectiveness and generalization of our method.

    % 我们还将我们的方法与其他text-to-SQL纠错方法进行比较，如表所示
    We also compare our method with other text-to-SQL correction methods, as shown in Table~\ref{tab:comparison_with_other_system}. 
    % 可以发现，在gpt-3.5-turbo上，我们的方法在基线性能更低的情况下，取得了更好的性能提升，证明了我们的方法在gpt-3.5-turbo上要优于现有的text-to-SQL纠错方法
    It can be observed that on GPT-3.5, our method achieves better performance improvement despite having a lower baseline performance, demonstrating that our method outperforms existing text-to-SQL correction methods on GPT-3.5.

    % 从表中，我们还可以发现：
    Besides, from Table~\ref{tab:main_experiment}, we can also see that:

    \paragraph{Dataset}
        % 在不同的数据集上，我们的方法都带来了性能提升
        \ourmethod improves performance across different datasets.
        % 相较于Spider，我们的方法在Bird和KaggleDBQA上的性能提升更加显著
        Compared to Spider, our method achieves more performance gains on Bird and KaggleDBQA in most settings.
        % 这是因为对于更复杂的问题和数据库，模型更有可能产生实体连接错误，或者生成错误的骨架
        This is because, with more complex questions and databases, LLMs are more likely to make entity-linking mistakes or generate incorrect skeletons.
        % 因此，我们的方法能更好地提升模型性能
        Consequently, our method enhances model performance more effectively in Bird and KaggleDBQA by correcting skeleton or entity mistakes.

    \paragraph{Model}
        % 在不同的模型上，我们的方法也同样带来了显著的性能提升
        Our method also brings significant performance improvements across different models. 
        % 相较于代码大模型，我们的方法在通用大模型上的提升更显著
        Compared to code-specific LLMs, the enhancement is more pronounced in general LLMs. 
        % 这是因为，代码大模型在代码数据上进行微调，他在非代码任务上的指令跟随能力要弱于通用大模型，因此在实体链接和骨架生成两个任务上的性能相对较弱
        This is because the code LLMs are fine-tuned on the code data, making its instruction following ability on non-code tasks weaker than that of the general LLMs, leading to poorer performance in entity linking and skeleton parsing.

    \paragraph{Scale}
        % 对于不同规模规模的模型，我们的方法均带来了性能提升
        \ourmethod improves performance across models of different scales.
        % 相较于large-scale的模型，我们的方法在small-scale模型上带来的性能提升更显著
        The performance improvement is more significant for small-scale models compared to large-scale models.
        % 这是因为，small-scale的模型由于性能限制，更容易在实体、骨架上产生错误
        This is because small-scale models are more prone to mistakes in entities and skeletons due to their performance limitations. 
        % 而我们的方法通过从子任务的角度进行纠错，能有效地引导模型生成正确的结果
        Our method, by correcting from the perspective of sub-tasks, effectively guides the model to generate correct results.

\subsection{Ablation Study}
    \begin{table*}[t]
        \small
        \centering
        \begin{tabular}{lll|ccc}
    \toprule
    \textbf{Model} & \textbf{Scale} & \textbf{Method} & \textbf{Spider} & \textbf{Bird} & \textbf{Kaggle} \\
    \midrule
    \multirow{6}{*}{Llama3} & \multirow{3}{*}{8b} & \ourmethod & $75.0$ & $37.3$ & $29.4$ \\
     & & \quad - Entity & $74.9 (-0.1)$ & $36.4 (-0.9)$ & $29.2 (-0.2)$ \\
     & & \quad - Skeleton & $73.3 (-1.7)$ & $35.3 (-2.0)$ & $29.0 (-0.4)$ \\
    \cmidrule{2-6}
     & \multirow{3}{*}{70b} & \ourmethod & $81.3$ & $48.4$ & $35.1$ \\
     & & \quad - Entity & $81.2 (-0.1)$ & $47.1 (-1.3)$ & $34.9 (-0.2)$ \\
     & & \quad - Skeleton & $80.9 (-0.4)$ & $47.1 (-1.3)$ & $33.8 (-1.3)$ \\
    \midrule
    \multirow{6}{*}{Deep.C.} & \multirow{3}{*}{6.7b} & \ourmethod & $77.1$ & $44.7$ & $25.4$ \\
     & & \quad - Entity & $77.0 (-0.1)$ & $44.1 (-0.6)$ & $25.4 (-0.0)$ \\
     & & \quad - Skeleton & $76.5 (-0.6)$ & $44.6 (-0.1)$ & $25.2 (-0.2)$ \\
    \cmidrule{2-6}
     & \multirow{3}{*}{33b} & \ourmethod & $80.1$ & $52.7$ & $29.2$ \\
     & & \quad - Entity & $79.9 (-0.2)$ & $52.1 (-0.6)$ & $28.3 (-0.9)$ \\
     & & \quad - Skeleton & $79.4 (-0.7)$ & $51.6 (-1.1)$ & $28.9 (-0.3)$ \\
    \bottomrule
\end{tabular}

        \caption{
            The ablation experiments of \ourmethod on:
            \textit{(i) Entity}: remove the linked entities during the correction;
            \textit{(ii) Skeleton}: remove the parsed skeleton during the correction.
        }
        \label{tab:ablation_study}
    \end{table*}

    % 为了验证我们方法设计的各个部分的正确性，我们对实体连接和骨架生成两部分进行了消融实验，实验结果如表所示
    To validate the effectiveness of each step that our method designed, we conduct ablation experiments on entity linking and skeleton parsing. 
    % 从表中我们可以发现：
    The results are shown in Table~\ref{tab:ablation_study}, from which we can see that:
    % (i) 在删除实体连接和骨架生成后，我们的方法都产生了性能下降，证明了我们设计的方法各个部分的有效性
    \textit{(i)} Removing either entity linking or skeleton parsing results in a performance drop, confirming the effectiveness of each step of our design;
    % (ii) 相较于实体连接，删除骨架生成后带来的性能下降更显著，证明现有模型的性能瓶颈主要在骨架生成上
    \textit{(ii)} The performance drop is more significant when skeleton parsing is removed compared to entity linking, indicating that the primary performance bottleneck of the current model lies in skeleton parsing;
    % (iii) 相对于规模更大的模型，小模型上模块消融后的性能下降更显著，表明小模型生成的结果的错误更加显著，自纠错能带来更大的性能提升
    \textit{(iii)} The performance degradation is more pronounced in models with smaller scales after ablation, suggesting that errors in the results generated by smaller models are more significant, and thus \ourmethod can bring greater performance improvements.

\subsection{Analysis}
    % 拥有正确实体和骨架的SQL是不是正确的SQL？
    \subsubsection{Is SQL with the correct entity and skeleton the correct SQL?}
        \begin{table}[t]
            \centering
            \small
            \begin{tabular}{ll|ccc}
    \toprule
    \textbf{Model} & \textbf{Scale} & \textbf{Spider} & \textbf{Bird} & \textbf{Kaggle} \\
    \midrule
    \multirow{2}{*}{Llama3} & 8b & $99.7$ & $93.8$ & $94.0$ \\
     & 70b & $99.7$ & $93.3$ & $96.8$ \\
    \midrule
    \multirow{2}{*}{Deep.C.} & 6.7b & $99.7$ & $96.1$ & $98.9$ \\
     & 33b & $99.7$ & $94.2$ & $96.0$ \\
    \bottomrule
\end{tabular}

            \caption{
                % 不同模型和数据集上的结果准确率，计算方式为，所有实体和骨架与答案完全相同的结果中，正确的答案的比例
                The proportion of the correct SQL in the results that the linked entities and the skeleton are correct.
            }
            \label{tab:result_accuracy}
        \end{table}

        % 在Discussion中我们提出，从子任务的角度进行纠错得到的结果可能不是正确的SQL
        In the Discussion section, we propose that automated correction from the perspective of sub-tasks could not yield correct SQL results.
        % 为了论证我们方法的有效性，我们统计了实体和骨架正确的SQL中，正确的SQL的比例，如表所示
        To demonstrate the effectiveness of \ourmethod, we calculate the proportion of correct SQL among those with correct entities and skeletons, as shown in Table~\ref{tab:result_accuracy}. 
        % 从表中可以发现：
        From the table, we can observe that:
        % (i) 大部分设置下的结果都接近100\%，证明了从实体和骨架的角度纠错正确和正确的SQL之间有很强的一致性，证明了我们方法的有效性
        \textit{(i)} The results under most settings are close to $100\%$, proving a strong consistency between automated correction from the perspectives of entities and skeletons and the correct SQL, thereby validating the effectiveness of our method;
        % (ii) 相较于Spider，Bird和KaggleDBQA上结果的准确性更低，表明这两个数据集难度更高，更需要实体连接和骨架生成之外的能力来生成正确的SQL
        \textit{(ii)} Compared to Spider, the accuracy of results on Bird and KaggleDBQA is lower, indicating that these two datasets are more challenging and require abilities beyond entity linking and skeleton parsing to produce correct SQL.

    % 大模型是否能有效地生成实体链接信息？
    \subsubsection{Can LLMs effectively generate entity linking information?}
        \begin{table}[t]
            \centering
            \small
            \begin{tabular}{l|ccc|ccc}
    \toprule
    \multirow{2}{*}{\textbf{Method}} & \multicolumn{3}{c|}{\textbf{Table}} & \multicolumn{3}{c}{\textbf{Column}} \\
    & \textbf{P} & \textbf{R} & \textbf{F} & \textbf{P} & \textbf{R} & \textbf{F} \\
    \midrule
    N-Gram & $$78.2$$ & $$69.6$$ & $$73.6$$ & $$61.4$$ & $$69.1$$ & $$65.1$$ \\
    EtA+BERT & $$81.1$$ & $$85.3$$ & $$83.1$$ & $$86.1$$ & $$79.3$$ & \bm{$82.5$} \\
    \midrule
    Llama3-8b & $$80.9$$ & $$86.4$$ & $$81.6$$ & $$76.6$$ & $$78.3$$ & $$75.5$$ \\
    Llama3-70b & \bm{$88.9$} & \bm{$89.1$} & \bm{$87.4$} & $$78.9$$ & \bm{$83.1$} & $$79.0$$ \\
    \midrule
    Deep.C.-6.7b & $$69.4$$ & $$74.6$$ & $$70.0$$ & $$72.8$$ & $$72.6$$ & $$70.3$$ \\
    Deep.C.-33b & $$79.1$$ & $$82.5$$ & $$79.1$$ & \bm{$86.2$} & $$78.6$$ & $$79.9$$ \\
    \bottomrule
\end{tabular}

            \caption{
                The entity linking performance of different models on Spider.
                P, R, and F represent the macro average of accuracy, recognition rate, and F1 respectively.
                The results of N-Gram and EtA+BERT are reported by \citet{EtA}
                The best performances are marked in \textbf{bold}.
            }
            \label{tab:entity_linking}
        \end{table}

        % 为了验证大模型做实体连接的性能，我们在\citep标注了实体连接信息的Spider数据集上进行了实验
        To validate the performance of LLMs in entity linking, we conduct experiments on the schema linking of the Spider dev set.
        % 实验结果如表所示，从表中可以发现：
        The experimental results are shown in Table~\ref{tab:entity_linking}, from which we can see that:
        % (i) 在大部分指标上，基于大模型的实体连接性能都要高于基于词段和微调方法的性能，证明了大模型在做实体连接任务上的有效性
        \textit{(i)} In most metrics, the performance of entity linking based on LLMs is superior to that based on grams or fine-tuning methods \cite{EtA}, demonstrating the effectiveness of LLMs in entity linking;
        % (ii) 相较于代码大模型，通用大模型在大部分指标上表现出了更好的实体连接性能，表明通用大模型更适合做实体连接任务
        \textit{(ii)} Compared to code-specific LLMs, general LLMs exhibit better entity-linking performance in most metrics, indicating that general-purpose LLMs are more suitable for the entity-linking task due to the better instruction following.

    % 不基于数据库生成的SQL骨架是否比基于数据库生成的SQL骨架更好
    \subsubsection{Is the SQL skeleton generated not based on databases better than based on databases?}
        \begin{table}[t]
            \centering
            \small
            \begin{tabular}{lllccc}
    \toprule
    \textbf{Model} & \textbf{Scale} & \textbf{Method} & \textbf{Spider} & \textbf{Bird} & \textbf{Kaggle} \\
    \midrule
    \multirow{4}{*}{Llama3} & \multirow{2}{*}{8b} & Initial & $32.0$ & $8.1$ & $18.2$ \\
     &  & Parsed & $\bm{35.1}$ & $\bm{10.1}$ & $\bm{19.5}$ \\
    \cmidrule{2-6}
     & \multirow{2}{*}{70b} & Initial & $36.1$ & $11.7$ & $24.3$ \\
     &  & Parsed & $\bm{37.4}$ & $\bm{12.0}$ & $\bm{25.2}$ \\
    \midrule
    \multirow{4}{*}{Deep.C.} & \multirow{2}{*}{6.7b} & Initial & $36.0$ & $13.2$ & $15.4$ \\
     &  & Parsed & $\bm{37.7}$ & $\bm{18.3}$ & $\bm{19.9}$ \\
    \cmidrule{2-6}
     & \multirow{2}{*}{33b} & Initial & $35.0$ & $11.5$ & $17.5$ \\
     &  & Parsed & $\bm{36.5}$ & $\bm{16.2}$ & $\bm{19.3}$ \\
    \bottomrule
\end{tabular}

            \caption{
                The skeleton accuracy of initial SQL and parsed skeletons.
                The best performance of each model and each dataset is marked in \textbf{bold}.
            }
            \label{tab:skeleton_hallucination}
        \end{table}

        % 为了验证大模型生成的骨架是否准确性比直接生成SQL的性能要高，我们验证了各个数据集和模型上生成的骨架的准确性
        To verify whether the performance of skeletons generated by LLMs is higher than directly generating SQL, we evaluate the accuracy of the skeletons generated on various datasets and models.
        % 实验结果如表所示，从表中我们可以发现：
        The experimental results are shown in Table~\ref{tab:skeleton_hallucination}, from which we can observe:
        % (i) 在所有数据集和所有模型上，生成的骨架的准确性比直接生成SQL的骨架的准确性要高，证明了我们使用生成骨架来纠错的有效性
        \textit{(i)} Across all datasets and models, the accuracy of the parsed skeletons is higher than that of directly generated SQL, demonstrating the effectiveness of using parsed skeletons for correction;
        % (ii) 相对于规模更大的模型，小规模模型上生成的骨架准确性的提升要更高，这是因为小规模模型鲁棒性要弱于大规模模型，受到数据库信息扰动的影响更严重
        \textit{(ii)} The improvement in skeleton accuracy is more significant in LLMs with smaller scales since small-scale models are less robust and more affected by disturbances in database information, leading to worse initial SQL;
        % (iii) 生成骨架的准确率要低于Table中生成SQL的准确率，这是因为相同的问题可能对应多种不同的SQL骨架，因此Table中的结果只能一定程度反应骨架与问题语意之间的相关程度
        \textit{(iii)} The accuracy of generating skeletons is lower than that of generating SQL in Table~\ref{tab:main_experiment} because one question could correspond to multiple SQL skeletons, so the results in Table~\ref{tab:skeleton_hallucination} can only reflect the correlation between the skeleton and the question to a certain extent.

    \subsubsection{What is the performance of \ourmethod using oracle entities and skeletons?}
        \begin{table*}[t]
            \centering
            \small
            \begin{tabular}{lll|ccc}
    \toprule
    \textbf{Model} & \textbf{Scale} & \textbf{Method} & \textbf{Spider} & \textbf{Bird} & \textbf{Kaggle} \\
    \midrule
    \multirow{8}{*}{Llama3} & \multirow{4}{*}{8b} & \ourmethod & $75.0$ & $37.3$ & $29.4$ \\
    & & \quad + Oracle Entity & $75.9 (+0.9)$ & $37.7 (+0.4)$ & $29.8 (+0.4)$ \\
    & & \quad + Oracle Skeleton & $75.6 (+0.6)$ & $37.9 (+0.6)$ & $30.5 (+1.1)$ \\
    & & \quad + Oracle Both & $76.4 (+1.4)$ & $38.0 (+0.7)$ & $30.7 (+1.3)$ \\
    \cmidrule{2-6}
    & \multirow{4}{*}{70b} & \ourmethod & $81.3$ & $48.4$ & $35.1$ \\
    & & \quad + Oracle Entity & $81.8 (+0.5)$ & $48.6 (+0.2)$ & $37.3 (+2.2)$ \\
    & & \quad + Oracle Skeleton & $83.6 (+2.3)$ & $48.9 (+0.5)$ & $36.9 (+1.8)$ \\
    & & \quad + Oracle Both & $83.8 (+2.5)$ & $49.5 (+1.1)$ & $38.6 (+3.5)$ \\
    \midrule
    \multirow{8}{*}{Deepssek-Coder} & \multirow{4}{*}{6.7b} & \ourmethod & $77.1$ & $44.7$ & $25.4$ \\
    & & \quad + Oracle Entity & $77.9 (+0.8)$ & $44.7 (+0.0)$ & $26.3 (+0.9)$ \\
    & & \quad + Oracle Skeleton & $78.9 (+1.8)$ & $44.8 (+0.1)$ & $26.5 (+1.1)$ \\
    & & \quad + Oracle Both & $78.9 (+1.8)$ & $45.4 (+0.7)$ & $27.2 (+1.8)$ \\
    \cmidrule{2-6}
    & \multirow{4}{*}{33b} & \ourmethod & $80.1$ & $52.7$ & $29.2$ \\
    & & \quad + Oracle Entity & $80.7 (+0.6)$ & $52.7 (+0.0)$ & $31.1 (+1.9)$ \\
    & & \quad + Oracle Skeleton & $81.0 (+0.9)$ & $52.8 (+0.1)$ & $30.0 (+0.8)$ \\
    & & \quad + Oracle Both & $81.6 (+1.5)$ & $53.1 (+0.4)$ & $31.1 (+1.9)$ \\
    \bottomrule
\end{tabular}

            \caption{
                The performance of \ourmethod with oracle entities and skeletons.
                Entity and Skeleton denote using the oracle entities and the oracle skeleton respectively, and Both denote using the oracle of them both.
            }
            \label{tab:oracle_experiment}
        \end{table*}

        % 为了探索我们方法性能提升的瓶颈，我们使用oracle的实体和骨架进行了实验
        To explore the bottlenecks in improving the performance of \ourmethod, we conduct experiments using oracle entities and skeletons.
        % 实验结果如表所示，从表中我们可以发现：
        The experimental results are shown in Table~\ref{tab:oracle_experiment}, from which we can observe:
        % (i) 在引入oracle的实体或骨架后，模型的性能有进一步的提升，证明了增强实体连接或骨架生成的性能能进一步增强模型性能
        \textit{(i)} Introducing oracle entities or skeletons further improves the performance of \ourmethod, demonstrating that enhancing entity linking or skeleton parsing can further enhance the performance;
        % (ii) 使用oracle骨架的性能提升要高于使用oracle实体，证明骨架生成的性能限制要比实体连接要严重
        \textit{(ii)} The performance improvement using oracle skeletons is greater than using oracle entities, indicating that the performance limitation of skeleton parsing is more severe than that of entity linking;
        % (iii) 使用oracle的性能提升并不显著，表明直接用大模型基于错误信息进行纠错的效果并不显著，未来工作需要探索更有效的方法来帮助模型将连接的实体或生成的骨架应用到生成的SQL上
        \textit{(iii)} The performance improvement using the oracle entities or oracle skeletons is not significant, suggesting that direct correction by LLMs is not effective, where future work should explore more effective methods to help LLMs align the SQL to the linked entities or generated skeletons.

    \subsubsection{What is the performance of \ourmethod on SQLs with different hardness?}
        \begin{table}[t]
            \centering
            \small
            \begin{tabular}{ll|cccc}
    \toprule
    \textbf{Model} & \textbf{Method} & \textbf{Easy} & \textbf{Medium} & \textbf{Hard} & \textbf{Extra} \\
    \midrule
    \multirow{2}{*}{Llama3-8b} & baseline & $87.1$ & $78.0$ & $59.8$ & $38.0$ \\
    & + \ourmethod & \bm{$89.9$} & \bm{$81.8$} & \bm{$61.5$} & \bm{$48.8$} \\
    \midrule
    \multirow{2}{*}{Llama3-70b} & baseline & $92.3$ & $87.0$ & $68.4$ & $59.0$ \\
    & + \ourmethod & \bm{$94.0$} & \bm{$86.5$} & \bm{$69.0$} & \bm{$61.4$} \\
    \midrule
    \multirow{2}{*}{Deep.C.-6.7b} & baseline & $86.7$ & $83.0$ & $61.5$ & $44.6$ \\
    & + \ourmethod & \bm{$90.3$} & \bm{$84.8$} & \bm{$66.7$} & \bm{$47.6$} \\
    \midrule
    \multirow{2}{*}{Deep.C.-33b} & baseline & $92.3$ & $85.7$ & $59.8$ & $48.8$ \\
    & + \ourmethod & \bm{$95.2$} & \bm{$86.8$} & \bm{$62.1$} & \bm{$58.4$} \\
    \bottomrule
\end{tabular}

            \caption{
                Performance on questions with different hardness of the Spider dev set with and without \ourmethod.
                The best performance of different settings is marked in \textbf{bold}.
            }
            \label{tab:hardness}
        \end{table}
    
        % 为了评测我们的方法在不同难度上的问题上的性能，我们对Spider上不同难度的问题的性能进行了实验
        To evaluate the performance of \ourmethod on questions of varying difficulty, we analyze the results of the Spider dev set, which is shown in Table~\ref{tab:hardness}.
        % 实验结果如表所示，从表中我们可以发现：
        From the table, we can observe:
        % (i) 在不同难度的问题上，我们的方法都带来了显著的性能提升，证明了我们方法的有效性
        \textit{(i)} Our method significantly improves performance across questions of different hardness, demonstrating its effectiveness on both easy and hard questions;
        % (ii) 相较于简单问题，我们的方法在难的问题上带来的性能提升更显著，这是因为在难的问题上，模型更有可能在实体连接或骨架生成上产生错误
        \textit{(ii)} Compared to easy questions, the performance improvement is more pronounced on harder questions, since LLMs are more likely to make mistakes in entity linking or skeleton parsing.

\subsection{Case Study}
    \begin{figure}[t]
        \centering
        \includegraphics[width=0.6\linewidth]{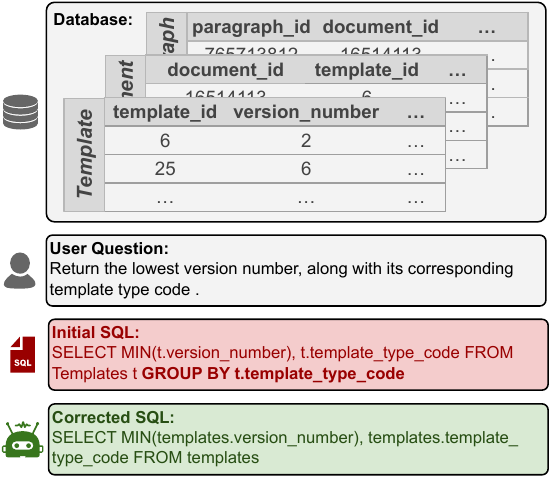}
        \caption{
            The case study with and without \ourmethod of the Spider dev set using Llama3-70b.
            The mistake in the SQL is annotated with \textbf{bold}.
        }
        \label{fig:case_study}
    \end{figure}

    % 尽管表中的结果已经证明了我们方法的有效性，然而我们的方法如何具体地增强text-to-SQL性能依然有待研究
    Despite the effectiveness of \ourmethod is demonstrated in Table~\ref{tab:main_experiment}, how our method specifically enhances text-to-SQL performance remains to be studied.
    % 因此，我们进行了case study，如图所示
    Therefore, we conduct a case study, as shown in Figure~\ref{fig:case_study}.
    % 在没有使用我们的方法时，模型生成的结果遗漏了使用实体，并且生成了错误的骨架
    From the figure, we can see that, without our method, the model output misses using entities and generates an incorrect skeleton.
    % 在使用我们的方法后，模型生成了正确的骨架，并且用到了所有问题相关的实体，从而增强了text-to-SQL性能
    With our method, the model generates the correct skeleton and utilizes all relevant entities, thereby enhancing text-to-SQL performance.

    \section{Related Works}
        \subsection{Text-to-SQL}
    % Text-to-SQL任务研究如何根据给定的用户问题和数据库来生成相应的SQL结果
    Research on the text-to-SQL task investigates how to generate corresponding SQL results based on given user questions and databases \citep{Text2SQL-Survey-PLM}.
    % 过去，人们研究主要基于小规模模型，例如使用BERT等encoder模型来判断SQL中对应的实体或关键词，或使用BART等encoder-decoder模型将text-to-SQL任务看作是自然语言到SQL的翻译任务，直接生成相应的SQL
    The past research is mainly based on small-scale models  \cite{guo2019towards,bogin2019global,wang2020rat,scholak2021picard,dou2022unisar} like BERT~\cite{devlin2019bert} and BART~\cite{lewis2020bart}.
    % 目前主流的text-to-SQL方法是基于大模型
    Currently, the methods based on LLMs have become the mainstream of the text-to-SQL task since their brilliant performance without fine-tuning \citep{Text2SQL-Survey-LLM}.
    % 一类方法通过微调大模型，来帮助大模型更好地应用到text-to-SQL任务上
    One type of approach is to fine-tune LLMs to help them better adapt to the text-to-SQL task by directly using the text-to-SQL data \citep{FinSQL,CodeS,DTS-SQL}, or merging the data of the tasks related to the text-to-SQL task (e.g., entity linking, knowledge generation) \cite{QDS-SQL,Knowledge-to-SQL}.
    % 另一类方法直接使用大模型few-shot推理，性能显著优于微调的方法而无需微调的开销
    Another type of approach uses LLMs by few-shot inference, achieving significantly better performance without the overhead of fine-tuning, like in-context learning \cite{DIN-SQL,DAIL-SQL,ODIS}, self-consistency \cite{Lever,APEL,lee2024mcssql}, and multi-agent \cite{MAC-SQL,xie2024deasql}.

    % 然而，现有的方法无法有效地判断生成SQL的正确性，限制了其在实际场景下的应用
    However, it is hard for the existing methods to effectively judge the correctness of generated SQL, limiting their practical application.
    % 因此，我们提出了我们的方法，判断实体连接和骨架生成两个子任务上的正确性，从而判断生成SQL的正确性
    Therefore, we propose our method, which automatically assesses the correctness based on the decomposed sub-tasks of entity linking and skeleton parsing, thereby correcting the mistakes in the generated SQL.

\subsection{Automated Correction}
    % Self-Debug方法在生成答案之后，首先判断生成结果的对错，如果判断答案错误则进行相应的改正
    The automated correction approach first determines the correctness of the generated result after producing an answer, where if the answer is incorrect, it makes the necessary corrections \citep{Self-Debug-Survey}.
    % 最主流的self-debug方法主要关注代码生成任务，例如在训练时使用反馈来修正错误，或者推理时模型判断结果的错误与否并加以修正
    The most mainstream automated correction methods primarily focus on the code generation task, such as using feedback during training to correct errors \citep{RLHF,Self-Motivated,CriticGPT} or having the model judge and correct errors after inference \citep{CodeT,Self-Debug,ProCo,zhang2024selfcontrast}.
    % 而在text-to-SQL任务上，自纠错更加关注任务本身的特点，例如根据数据库上的执行结果来修正，或者生成SQL相关的错误信息
    For the text-to-SQL task, automated correction methods pay more attention to the characteristics of the task itself, such as correcting based on execution results from the database \citep{TestCase,APEL,SQLCraft} or generating SQL-related error messages \citep{EPI-SQL,MAGIC}.

    % 然而，过去的工作表明，大模型很难直接判断出SQL的错误与否以及错在何处
    However, previous work shows that LLMs struggle to directly identify SQL mistakes \cite{Benchmarking,kamoi2024SelfCorrectionSurvey}.
    % 因此，我们提出我们的方法，来直接向大模型反馈实体连接和骨架生成上的错误之处，从而帮助模型纠正自己的错误
    Therefore, we propose \ourmethod to provide decomposed feedback from the sub-tasks of entity linking and skeleton parsing, thereby helping correct mistakes.

    \section{Conclusion}
        % 在本文中，我们讨论了text-to-SQL任务中，大模型直接对生成的SQL进行纠错的效果较差的问题
        In this paper, we try to alleviate the problem that LLMs perform poorly when directly correcting generated SQL for the text-to-SQL task.
        % 我们首先提出了，将原始任务分解为子任务，针对子任务纠错的效果要好于直接对原始任务的结果进行纠错
        We first propose that correcting by decomposing the original task into sub-tasks is more effective than directly correcting the generated results. 
        % 基于上述结论，我们提出将text-to-SQL任务分解为实体链接和骨架生成两个子任务
        Based on this discussion, we propose \ourmethod that decomposes the text-to-SQL task into two sub-tasks: entity linking and skeleton parsing. 
        % 我们让大模型生成实体和骨架信息，判断与初始SQL的不一致之处作为反馈，来进行纠错
        We ask LLMs to generate entity and skeleton information, identify inconsistencies with the initial SQL as feedback, and use this feedback for correction. 
        % 实验结果表明，我们在三个主流text-to-SQL数据集上带来了3.7%的性能提升，证明了我们方法的有效性
        Experimental results show that our method brings a $3.7\%$ performance improvement on three mainstream text-to-SQL datasets, demonstrating the effectiveness of our method.

    \clearpage
    \bibliography{colm2024_conference}
    \bibliographystyle{colm2024_conference}
    \clearpage

    % \section{Reproducibility Checklist}
    %     \input{tex/6.checklist}
    % \clearpage

    \appendix
    \section{Prompts of \ourmethod}
    \label{app:prompts}

    \begin{table*}[t]
        \centering
        \small
        \begin{tabular}{p{0.9\textwidth}}
            \toprule
            \textbf{The prompt of SQL Generation.} \\
            \midrule
            Generate a SQL to answer the question with the given schema. \\
            Quote your answer with: \\
            ```sql \\
            $<$answer sql$>$ \\
            ``` \\
            \\
            --- \\
            \\
            For example: \\
            \\
            ```sql \\
            \{schema of demonstrations\} \\
            ``` \\
            \\
            Question: \{question of demonstrations\} \\
            ```sql \\
            \{sql of demonstrations\} \\
            ``` \\
            \\
            --- \\
            \\
            ... \\
            \\
            --- \\
            \\
            Based on the instruction and the examples, answer the following question:  \\
            \\
            ```sql \\
            \{schema\} \\
            ``` \\
            \\
            Question: \{question\} \\
            \bottomrule
        \end{tabular}
        \caption{The prompt of SQL Generation of \ourmethod.}
        \label{tab:prompt_sql_generation}
    \end{table*}

    \begin{table*}[t]
        \centering
        \small
        \begin{tabular}{p{0.9\textwidth}}
            \toprule
            \textbf{The prompt of Entity Linking.} \\
            \midrule
            Align the tokens in the given question to the table entities or the column entities of the schema above, considering the given SQL. \\
            Present the aligned tokens in the python format List[Dict[str, str]], where each Dict[str, str] denoting each token in the question containing the following keys: \\
            \{ \\
                "token": the token in the question \\
                "schema": the schema entity aligned to the token \\
                "type": the type of the entity aligned to the token \\
            \} \\
            The "type" can be one of the following: \\
            * "tbl": the table name \\
            * "col": the column name \\
            * "val": the value \\
            "schema" and "type" are either both null or not null at the same time. \\
            \\
            Here are some examples. \\
            \\
            --- \\
            \\
            \{schema of demonstrations\} \\
            \\
            SQL: \{sql of demonstrations\} \\
            Question: \{question of demonstrations\} \\
            Alignments: \{alignment of demonstrations\} \\
            \\
            --- \\
            \\
            ... \\
            \\
            --- \\
            \\
            Based on the instruction and the examples above, solve the following question: \\
            \\
            \{schema\} \\
            \\
            SQL: \{sql\} \\
            Question: \{question\} \\
            Alignments: \\
            \bottomrule
        \end{tabular}
        \caption{The prompt of Entity Linking of \ourmethod.}
        \label{tab:prompt_entity_linking}
    \end{table*}
    
    \begin{table*}[t]
        \centering
        \small
        \begin{tabular}{p{0.9\textwidth}}
            \toprule
            \textbf{The prompt of Skeleton Parsing.} \\
            \midrule
            Hallucinate a SQL to answer the question. \\
            Quote your answer with: \\
            ```sql \\
            $<$answer sql$>$ \\
            ``` \\
            \\
            --- \\
            \\
            For example: \\
            \\
            Question: \{question of demonstrations\} \\
            ```sql \\
            \{sql of demonstrations\} \\
            ``` \\
            \\
            --- \\
            \\
            ... \\
            \\
            --- \\
            \\
            Based on the instruction and the examples, answer the following question: \\
            \\
            Question: \{question\} \\
            \bottomrule
        \end{tabular}
        \caption{The prompt of Skeleton Parsing of \ourmethod.}
        \label{tab:prompt_skeleton_hallucination}
    \end{table*}
    
    \begin{table*}[t]
        \centering
        \small
        \begin{tabular}{p{0.9\textwidth}}
            \toprule
            \textbf{The prompt of Correction with the entity feedback.} \\
            \midrule
            ```sql \\
            \{schema\} \\
            ``` \\
            \\
            Fix the sql "\{sql\}" to answer the question "\{question\}" based on the above database and the alignment. \\
            Present your sql in the format: \\
            ```sql \\
            $<$your sql$>$ \\
            ``` \\
            It should be noticed that \{notification\}. Your sql must contain the tables and columns mentioned by the question. \\
            \bottomrule
        \end{tabular}
        \caption{
            The prompt of Correction of \ourmethod with the entity feedback.
            The format of ``\{notification\}'' is like ``$<$tables or columns$>$ are mentioned by the question''.
        }
        \label{tab:prompt_debug_entities}
    \end{table*}

    \begin{table*}[t]
        \centering
        \small
        \begin{tabular}{p{0.9\textwidth}}
            \toprule
            \textbf{The prompt of Correction with the skeleton feedback.} \\
            \midrule
            ```sql \\
            \{schema\} \\
            ``` \\
            \\
            Fix the sql "\{sql\}" to answer the question "\{question\}" with the above schema. \\
            Present your sql in the format: \\
            ```sql \\
            $<$your sql$>$ \\
            ``` \\
            It should be noticed that the SQL skeleton could be like "\{skeleton\}", where each '\_' can only be replaced with one single table, column or value. \\
            \bottomrule
        \end{tabular}
        \caption{
            The prompt of Correction of \ourmethod with the skeleton feedback.
        }
        \label{tab:prompt_debug_skeletons}
    \end{table*}

    The prompts of \ourmethod is shown in Table~\ref{tab:prompt_sql_generation}, Table~\ref{tab:prompt_entity_linking}, Table~\ref{tab:prompt_skeleton_hallucination}, Table~\ref{tab:prompt_debug_entities}, and Table~\ref{tab:prompt_debug_skeletons}..

\section{Error Analysis}
    \label{app:error_analysis}

    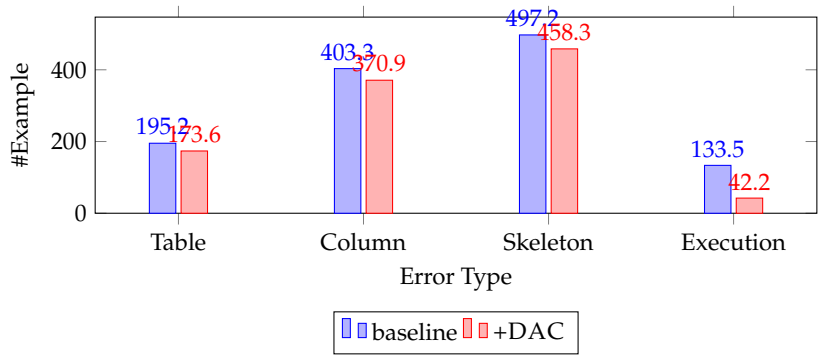
\begin{figure*}[t]
        \small
        \centering
        \begin{tikzpicture}
    \begin{axis}[
        ybar,
        symbolic x coords={Table, Column, Skeleton, Execution},
        xtick=data,
        nodes near coords,
        ymin=0,
        bar width=10pt,
        enlarge x limits=0.15,
        legend style={at={(0.5,-0.5)}, anchor=north,legend columns=-1},
        ylabel={\#Example},
        xlabel={Error Type},
        width=0.8\textwidth,
        height=0.3\textwidth
    ]
        \addplot coordinates {(Table,195.2) (Column,403.3) (Skeleton,497.2) (Execution,133.5)};
        \addplot coordinates {(Table,173.6) (Column,370.9) (Skeleton,458.3) (Execution,42.2)};
        \legend{baseline,+\ourmethod}
    \end{axis}
\end{tikzpicture}
        \caption{
            The error analysis with and without \ourmethod.
            \#Example denotes the average error examples across both models (Llama3, Deepseek-Coder) and all three datasets (Spider, Bird and KaggleDBQA).
            The table and column errors denote that the generated SQL does not have corresponding tables and columns of the correct SQL.
            The skeleton error denotes not having the correct skeleton.
            The execution error denotes that the generated SQL is not able to be executed.
            There could be one example that has multiple errors.
        }
        \label{fig:error_analysis}
    \end{figure*}

    % 为了分析我们方法的不足，启发未来的研究，我们统计了我们方法的错例
    To analyze the limitations of \ourmethod and inspire future research, we compile statistics on the errors produced by our method.
    % 我们按照表错误、列错误、骨架错误和执行错误来对错例进行分类
    We categorize these errors into \textit{table errors}, \textit{column errors}, \textit{skeleton errors}, and \textit{execution errors}.
    % 具体的统计结果如图所示，从图中我们可以发现：
    The specific statistical results are shown in Figure~\ref{fig:error_analysis}, from which we can observe:
    % (i) 在各类错误上，我们的方法都带来了显著的错误下降，证明了我们方法的有效性
    \textit{(i)} Our method significantly reduces errors across all categories, demonstrating its effectiveness;
    % (ii) 目前最主要的错误依然是骨架生成，这可能是因为大模型预训练时见过的常见的SQL骨架和数据集标注的骨架类型存在显著的差异
    \textit{(ii)} The primary error remains in skeleton generation due to that one user question could correspond to multiple different SQLs;
    % (iii) 除了骨架错误，列错误也是一个显著的错误，表明如何有效地引导模型检索并使用问题相关的列名依然是有待解决的问题
    (iii) Besides skeleton errors, column errors are also significant, indicating that effectively guiding LLMs to retrieve and use relevant columns remains under discovery.

\end{document}